\pdfoutput=1
\documentclass[11pt, twocolumn]{article}
\usepackage{multirow}
\usepackage{hyperref}
\usepackage{makecell}
\usepackage[numbers]{natbib}

\usepackage{graphicx}  
\usepackage{float}    

\usepackage{amsmath}
\usepackage[margin=1in, top=1in]{geometry}
\usepackage{caption}
\usepackage{times}
\usepackage{latexsym}
\usepackage[T1]{fontenc}

\usepackage[utf8]{inputenc}
\usepackage{microtype}
\usepackage{inconsolata}
\usepackage{graphicx}

\usepackage{comment}
\title{\textbf{Human Bias in the Face of AI: Examining Human Judgment Against Text Labeled as AI Generated}}

\author{
  \textbf{Tiffany Zhu\textsuperscript{1}},
  \textbf{Iain Weissburg\textsuperscript{2}}, 
  \textbf{Kexun Zhang\textsuperscript{3}},
  \textbf{William Yang Wang\textsuperscript{2}}, \\
  \textsuperscript{1}The Harker School,
  \textsuperscript{2}University of California, Santa Barbara, 
  \textsuperscript{3}Carnegie Mellon University,
  \\
  \small{
    \textbf{Correspondence:} \href{mailto:26tiffanyz@gmail.com}{26tiffanyz@gmail.com}
  }
}

\begin{document}
\vspace{-4cm} 

\date{}

\maketitle

\begin{abstract}
As AI advances in text generation, human trust in AI generated content remains constrained by biases that go beyond concerns of accuracy. This study explores how bias shapes the perception of AI versus human generated content. Through three experiments involving text rephrasing, news article summarization, and persuasive writing, we investigated how human raters respond to labeled and unlabeled content. While the raters could not differentiate the two types of texts in the blind test, they overwhelmingly favored content labeled as ``Human Generated,'' over those labeled ``AI Generated,'' by a preference score of over 30\%. We observed the same pattern even when the labels were deliberately swapped. This human bias against AI has broader societal and cognitive implications, as it undervalues AI performance. This study highlights the limitations of human judgment in interacting with AI and offers a foundation for improving human-AI collaboration, especially in creative fields.
\end{abstract}

\section{Introduction}

With the increasing accessibility of large language models (LLMs), such as OpenAI's ChatGPT \cite{chat}, Meta's LLaMA \cite{llama}, and Anthropic's Claude \cite{claude}, generative artificial intelligence(AI)'s capabilities are being utilized in expansive tasks. Their sophisticated text generation abilities raise both excitement and concern about the potential displacement of human workers \cite{trivedi}.

\begin{figure}[!h] 
  \includegraphics[width=\columnwidth]{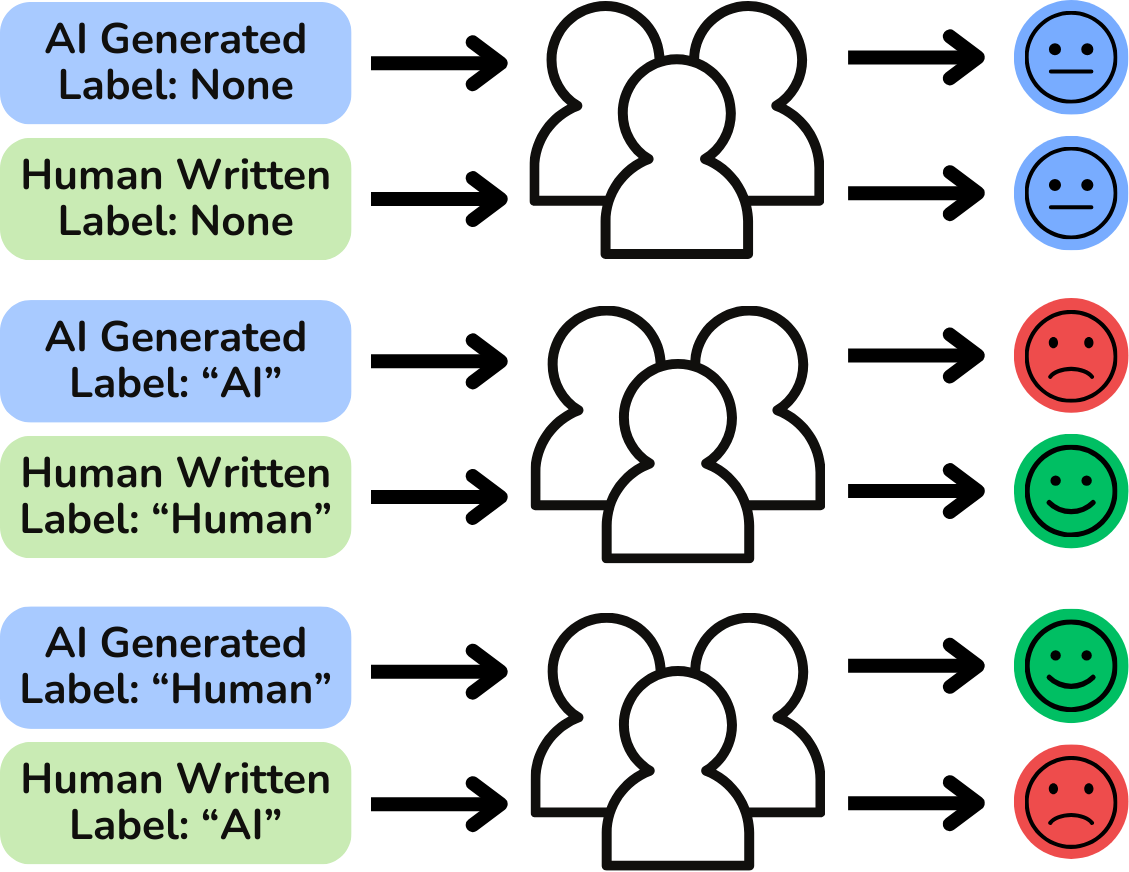} 
  \vspace{-1.5em}  
  \caption{Without labels for the texts, raters have little preference for AI generated or human written texts. With labels, raters prefer texts labeled ``Human Generated'' even when the labels were purposely swapped.}
  \label{fig:teaser}
\end{figure}

Public confidence in AI is critical for its successful integration into society. Positive perception of AI generated content helps forge a collaborative and productive relationship between humans and AI. Without this trust, the use of LLMs could face significant resistance, leading to underutilization \cite{choung}.

Prior research on AI mistrust has focused primarily on AI's bias towards different human populations, examining how biases can manifest in NLP models \cite{chang2017}. Specific studies discovered that AI systems can particularly amplify human biases regarding gender and race \cite{sun2019, sun2020}. While such work is important for understanding AI bias, our research introduces a different perspective: examining the bias humans have against AI.

Our work connects to interdisciplinary research on human-computer algorithm interaction. A study found that algorithm aversion depends on task type, with humans resisting AI in subjective tasks \cite{castelo2019}. Another earlier study showed people avoid algorithms after seeing them make even minor errors \cite{dietvorst2015}. More relevant to our approach, ``human favoritism'' was observed when evaluators knew content sources \cite{zhang2023}. Similarly, trust in humans versus ChatGPT was explored, focusing on explicit ratings \cite{buchanan2024}. Our study advances this through controlled blind tests and label-swapping experiments that more conclusively demonstrate bias against AI content.

Since their invention, humans have accepted computers for processing logical and mathematical tasks. ChatGPT 3.5's release in 2022 and subsequent AI improvements raised alarm regarding humans' perceived intellectual superiority in creativity and language processing \cite{gptresearch}. Past studies found that people view AI as inferior in work requiring originality \cite{creativity}. However, whether the belief that AI cannot match human skills in creative work like ``writing'' is rooted in reality or an expression of human bias requires further study.

This research aims to uncover whether biases distort judgments in writing. We focus on three scenarios that have common natural language processing applications: text rephrasing, where AI enhances the readability of existing text \cite{editing}; summarization, where AI condenses longer texts into concise forms \cite{summarization, summarization2}; and persuasion, where AI argues for a belief \cite{persuasion}.

We summarize our contributions in three folds:

\begin{itemize}
\vspace{-0.8em}
    \item This study investigated how human biases affect the perception of AI-generated text and whether these biases lead to assumptions that humans outperform AI in creative writing scenarios, an angle not explored before.
    \vspace{-0.8em}
    \item We designed a robust experimental framework comparing human and AI content that incorporates blind and manipulated labels to assess bias extent.
\vspace{-0.8em}
    \item Our findings, as previewed in Figure \ref{fig:teaser}, offer a basis for future research on human-AI collaboration, dataset creation, and public training to better understand AI capabilities.
\end{itemize}

\section{Methods}
\subsection{Data Collection}

Our data collection included three datasets available through Hugging Face \cite{huggingface}. The dataset \cite{wiki2} for text rephrasing contains human-written texts from Wikipedia article introductions. AI models were provided with human written introductions and asked to rephrase them for readability. The dataset for the summarization scenario \cite{see, hermann} contains full-length news articles from news sites like CNN and Daily Mail, alongside 2-3 sentence-long human written highlights. Provided with the news articles, AI models were instructed to write summaries. For the persuasion scenario, the dataset \cite{durmus} contains various controversial topics, such as ``Banning gas car sales too soon is unrealistic,'' and ``Social media should verify user identities.'' Each topic has a corresponding human written persuasive paragraph, and AI models were instructed to write persuasive articles with logical reasoning.

From each dataset, we selected 200 entries uniformly at random. We then sent each entry to three major LLMs accessed by API keys between August 10 and August 17, 2024 — Gpt-4o-2024-05-13 (ChatGPT), Claude-2 (Claude), and Meta-Llama-3.1-8B-Instruct (Llama) — to collect their responses. Using multiple models introduced a broader range of AI generated outputs while reducing the peculiarities of each model. In total, for each dataset, we collected 600 responses from the three AI models. The models were instructed to generate content with lengths similar to human-written content, as shown in Table \ref{tab:contentlen}.

\captionsetup[table]{skip=2pt} 
\begin{table}[h!]
  \centering
  \begin{tabular}{l@{\hskip 9pt}c@{\hskip 9pt}c}
    \hline
    \textbf{Experiment} & \textbf{Human} & \textbf{AI} \\
    \hline
    Text Rephrasing & 71 (SD 45) & 66 (SD 23) \\
    Summarization & 36 (SD 10) & 44 (SD 9) \\
    Persuasion & 256 (SD 42) & 231 (SD 27) \\
    \hline
  \end{tabular}
  \caption{Average word count and standard deviation (SD) comparison of human and AI generated contents.}
  \label{tab:contentlen}
\end{table}

\subsection{Experiment Design}

We used Amazon Mechanical Turk (MTurk) \cite{mturk} to gather the preferences of people from predominantly English-speaking nations (Australia, Canada, UK, US). MTurk was chosen because the demographic and socioeconomic profiles of the workers(taskers) there closely mirror the general public \cite{moss}. We required taskers to have >80\% HIT approval rate and manually evaluated their annotation quality. This threshold was used to achieve greater participant diversity while maintaining quality control.

For each scenario, specific guide questions were provided to help the taskers make appropriate evaluations. For text rephrasing, taskers were given a human written Wikipedia introductory paragraph and an AI rephrased version. The guiding question was: ``Which paragraph is better in terms of readability?'' \cite{pitler}. For summarization, taskers were provided with a full article and two summaries generated by human and AI, and the guiding question ``Which summary of the provided article presents the most relevant information in a clear manner?'' \cite{bischof}. For persuasion, taskers were presented with a topic and two persuasive texts, one generated by human and the other by AI, and ``Which persuasive paragraph supports the argument more?'' \cite{bizup}.

Our goal was to assess whether knowing the source of the generated texts influenced taskers' preferences when the content remained the same. We conducted two tests to identify people's preferences: a blind test and a manipulated test.

In the blind test, taskers were given the texts with randomly assigned labels ``1'' and ``2.'' They were asked to guess which text is more likely to be human generated, and then choose their preferences. We call this experiment Blind-Labeled or ``No Label'' and use it to check if taskers could tell the texts apart. If they cannot differentiate them, we then use the manipulated test to uncover biases.

The manipulated test contains two experiments: ``Correctly Labeled'' and ``Wrongly Labeled.'' In correctly labeled, the texts were explicitly and correctly labeled with ``Human Generated'' or ``AI Generated.'' The taskers were asked to choose their preference with the labels given. In wrongly labeled, taskers were given swapped labels. Unknown to the taskers, the AI generated text was labeled as ``Human Generated,'' and vice versa. They were then asked to choose their preferences.

Each experiment had 600 tasks (entries) to evaluate, and each task was assigned three taskers to work on. Overall, the total number of hits (evaluation results) we collected was 3 scenarios × 3 experiments/scenario × 600 tasks/experiment × 3 hits/task = 16200. Each tasker could evaluate up to three tasks per experiment but was blocked from working in multiple experiments for the same scenario. To ensure work quality, we excluded responses finished in less than 20 seconds (while still compensating the taskers), and reassigned the tasks to others. The eventual median completion times and compensation for rephrasing, summarization, and persuasion tasks were 3m 6s, \$0.42; 4m 4s, \$0.60; and 4m 9s, \$0.60, respectively. Appendix~\ref{sec:appendixa} shows an example of the MTurk task interface.

This study qualifies as Exempt from the University of California, Santa Barbara Human Subjects Committee IRB review under Category 3: Benign Behavioral Interventions. MTurk taskers received above federal minimum-wage compensation and were informed their anonymous hits would be used for data analysis.

\subsection{Data Analysis}

The survey result data were grouped by scenario, experiment, and LLM. We first assign scores to labeling choices for each task. Each choice is worth one point. For example, if two of the three taskers chose ``Human Generated'' for a task, that label receives two points, and the other ``AI Generated'' label receives one. We then took the average of the scores across all 600 tasks as the scores of an experiment. An average score of ``1.8'' for ``Human Generated'' label for an experiment means that 60\% of the taskers prefer ``Human-Generated'' texts in the experiment.

Our main goal was to study user preference change when presented texts with different labels. Hence, our findings reported in the next section will be centered around the labeling conditions. To study whether the preference changes across labeling conditions were significant, we report p-values from the Brunner-Munzel test \cite{brunner}.

\section{Results}

\paragraph{Taskers fail to differentiate between human and AI generated texts.} In the blind-labeled experiments, taskers were asked to guess which text was written by human. We found that, for the text rephrasing scenario, 49.93\% taskers responded incorrectly, i.e. they believed the AI rephrased texts were human generated. For the summarization scenario, the average was lower at 43.1\% responding incorrectly, but in the persuasion scenario, we again got a near-equal split of 50.06\%.

\paragraph{In blind tests, taskers show significant bias.} We conducted a chi-squared test \cite{chi} to identify if taskers have bias toward the text they prefer with four categories (text 1 is labeled AI and preferred; text 2 is labeled AI and preferred; 1 is labeled AI and 2 is preferred; and vice versa). Our chi-squared score was 28.4 with a p-value <.00001, so we determined there was bias among which text people prefer when no label is given.

Across all scenarios, taskers slightly preferred AI-generated text in blind-labeled experiments, with AI texts scoring above the 1.50 midpoint in all cases—1.537 for rephrasing, 1.650 for summarization, and 1.534 for persuasion.

\paragraph{Taskers prefer text labeled ``Human Generated'' even when wrongly classified.} Table \ref{tab:combined} reports the detailed score data for human written texts. We also broke down the scores by LLMs to show that the preferences over ``Human Generated'' texts hold across the board. We report percentage points instead of actual scores. The two can be converted to each other very easily. A value of 48.8\% in the table means a score of 3 * 0.488 = 1.464 for texts with ``Human-Generated'' labels.

Compared to blind tests, scores for human-generated texts in correctly labeled experiments increased by 32.9\% (rephrasing), 35.1\% (summarization), and 26.1\% (persuasion), showing that the taskers strongly prefer ``Human Generated'' texts. In the wrongly labeled experiments, the scores for the texts labeled ``Human Generated,'' but are actually ``AI Generated,'' also increased compared to the blind-labeled cases. For the text rephrasing scenario, the score for AI texts increased from 1.537 (51.2\%) to 2.052 (68.4\%) when they were labeled as ``Human Generated.'' Consequently, the score for human texts dropped from 1.463 (48.8\%) to 0.948 (31.6\%). Similar results happened for the summarization and persuasion scenarios as well, with drops of 47.8\% and 36.8\%, respectively.

Appendix~\ref{sec:appendixb} shows an example task from each scenario where human bias played a large role and all three taskers preferred the human text when correctly labeled as ``Human Generated'' and the AI text when wrongly labeled as ``Human Generated.''

\begin{table}[h!]
  \centering
  \begin{tabular}{l@{\hskip -11pt}l@{\hskip 3pt}c@{\hskip 3pt}c@{\hskip 3pt}c}
    \hline
    \textbf{Label} & \textbf{All} & \textbf{ChatGPT} & \textbf{Claude} & \textbf{Llama} \\
    \hline
    \textbf{Text Rephrasing} \\
    
      Blind    & 48.8\% & 47.0\% & 52.2\% & 47.2\% \\
      Correct & 64.8\% & 61.7\% & 66.5\% & 66.3\% \\
      Wrong   & 31.6\% & 31.3\% & 32.3\% & 31.2\% \\
    \hline
    \textbf{Summarization} \\
      Blind & 45.0\% & 46.3\% & 44.8\% & 43.8\% \\
    Correct & 60.8\% & 60.3\% & 59.2\% & 62.8\% \\
    Wrong & 23.5\% & 22.5\% & 23.0\% & 25.0\% \\
    \hline
    \textbf{Persuasion} \\
     Blind & 48.9\% & 49.4\% & 47.3\% & 49.8\% \\
    Correct & 63.7\% & 62.5\% & 65.7\% & 62.8\% \\
    Wrong  & 31.6\% & 32.2\% & 31.2\% & 31.4\% \\
    \hline
  \end{tabular}
  
  \caption{Proportion of taskers who preferred true human text. Across 3 scenarios and models, the proportion increased with ``Human Generated'' label and decreased with ``AI Generated'' label.}
  \label{tab:combined}
\end{table}

The Brunner-Munzel test confirmed significant differences between all labeling conditions (p<0.05) \cite{brunner}. Table \ref{tab:tests} presents all p-values.

\begin{table}
  \centering
  \begin{tabular}{l@{\hskip 9pt}c@{\hskip 9pt}c}
    \hline
    \textbf{Scenario} & \textbf{Label Conditions} & \textbf{p-value} \\
    \hline
    \multirow{3}{*}{\makecell{\textbf{Text} \\ \textbf{Rephrasing}}} & None \& Correct & 2.43E-18 \\
     & None \& Wrong & 7.90E-26 \\
     & Correct \& Wrong & 2.79E-93 \\
    \hline
    \multirow{3}{*}{\makecell{\textbf{Summari-} \\ \textbf{zation}}} & None \& Correct & 8.44E-23 \\
     & None \& Wrong & 9.25E-26 \\
     & Correct \& Wrong & 1.01E-97 \\
    \hline
    \multirow{3}{*}{\makecell{\textbf{Persuasion}}} & None \& Correct & 3.22E-22 \\
     & None \& Wrong & 4.23E-45 \\
     & Correct \& Wrong & 9.89E-136 \\
    \hline
  \end{tabular}
  \caption{Brunner-Munzel test statistics and p-values for comparisons of labeling conditions (blind labels, correct, wrong) across different scenarios.}
  \label{tab:tests}
\end{table}

\section{Discussion and Conclusion}

In our tests, taskers showed nearly equal ability in distinguishing between human and AI-generated text, consistent with prior research finding that linguists could only correctly identify AI-generated academic writing 38.9\% of the time \cite{casal}. Our results extend these findings to three other scenarios as well, suggesting that the challenge of differentiation remains true for all types of writing.

Our study also found that, when explicitly labeled, people prefer ``Human Generated'' texts more than ``AI Generated.'' One possible reason for the preference change is that people may mistrust LLMs due to their tendency to ``hallucinate,'' outputting inaccurate or misleading information. Although this might be the case for the persuasion scenario, where LLMs have greater generative creativity for argumentative writing \cite{Breum, pauli}, hallucinations are unlikely to occur in the rephrasing and summarization scenarios. For all scenarios, we manually examined a subset of output samples from each LLM and did not notice hallucinations. This finding, supported by the significant p-values, provides preliminary evidence that human bias against AI persists regardless of task complexity.

These biases hinder AI deployment and could impact AI alignment in systems that use Reinforcement Learning from Human Feedback (RLHF). RHLF uses human feedback to refine models like InstructGPT \cite{instructgpt}, but if humans favor content perceived as human-generated, AI systems may be trained to produce content aligning with these biased expectations.

For practitioners implementing NLP systems, we propose a few solutions to increase human trust in AI:

1. Creating transparent collaboration interfaces that position AI as an assistive tool rather than a competitor \cite{vossing}. Such interfaces can display AI reasoning processes and contributions in co-created content, helping users understand and fairly evaluate AI-generated text.

2. Implementing incremental AI integration strategies to build trust before expanding to critical applications \cite{solaiman}. Past research has demonstrated that user trust in technology evolves over time, suggesting that staged deployment across professional domains allows users to build positive relationships with AI systems gradually \cite{wu, cabiddu}.

3. Developing metrics for measuring human-AI collaborative effectiveness, providing objective data on AI's value that can help counter subjective human bias \cite{zerilli}.

4. Designing public education initiatives that accurately represent AI capabilities and limitations, combating inherent biases through increased understanding.

While our research identified human bias as a major factor against AI content, we still need to pinpoint the source of human bias. We suspect ``human ego'' plays a role. Humans are not ready to give up creative and intellectual superiority to AI, and therefore do not view AI content favorably. This hypothesis requires further investigation.

\section{Limitations}

One limitation in this study is that the data is collected from MTurk workers. Although the demographic is similar to the English-speaking public, specific biases may be inherent in this group, like their motivation for completing the hits.

A second limitation was that only three scenarios were selected. Other creative domains, like story writing and problem solving, can be further explored to gain a fuller understanding of the role of human bias in people's acceptance of AI.

\section{Ethical Considerations}

The ``Wrongly Labeled'' condition, where AI generated content was labeled as ``Human Generated'' and vice versa, does not pose ethical concerns for several reasons. First, this type of intervention is considered minimal risk in social science research since it does not expose participants to any physical, psychological, or emotional harm. The intervention is brief and necessary to understand how labeling conditions affects perceptions of text quality. This could not be accurately measured if participants were aware of the text origins.

Furthermore, the potential benefits of this research outweigh the minimal risks. By uncovering biases in how people perceive AI generated content, this study contributes valuable insights for the development and integration of AI into various fields. Understanding these biases is important for addressing potential barriers to AI adoption. Additionally, it could help develop strategies to mitigate unwarranted prejudices against AI generated content. This study's findings can lead to more effective human-AI collaboration and better decision-making processes in contexts where AI-generated content is increasingly prevalent. Therefore, the temporary and harmless intervention used in this study is justified by the significant potential benefits to the advancement of our understanding of human-AI interactions.

\bibliographystyle{plain}
\bibliography{custom}

\appendix

\section{Appendix: MTurk Interface}
\label{sec:appendixa}

\begin{figure}[H]
  \includegraphics[width=\columnwidth]{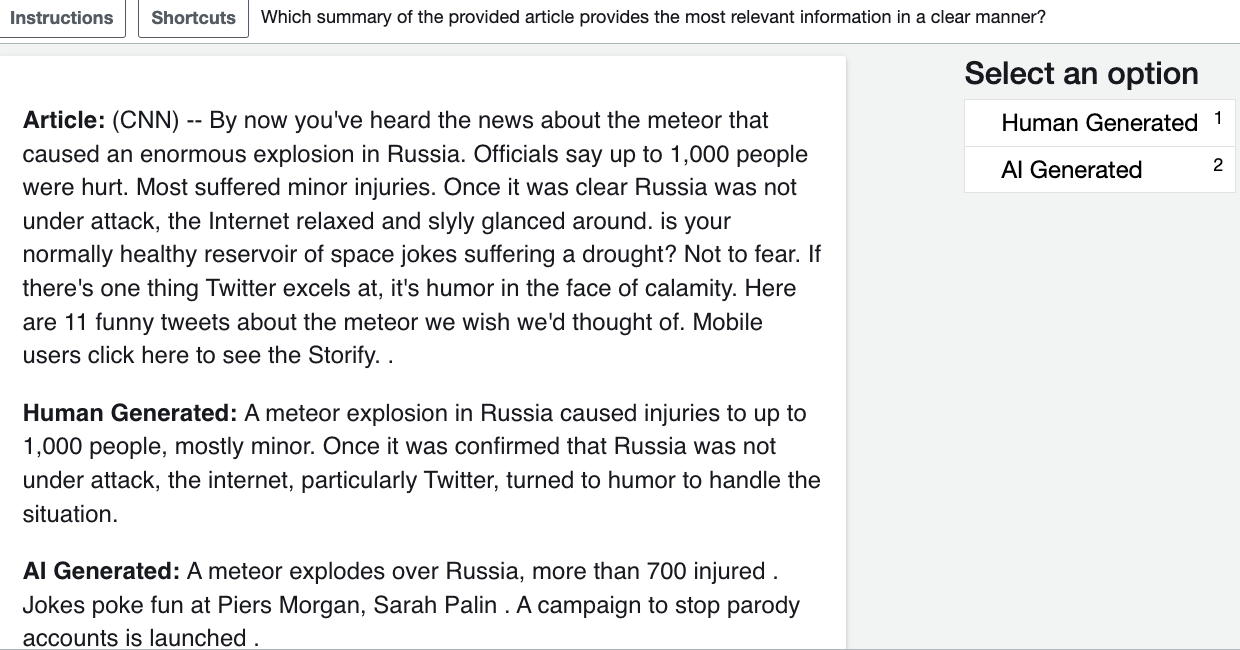}
    \vspace{-1.7em} 

  \caption{MTurk user interface for a correctly labeled summarization task.}
  \label{fig:experiments}
\end{figure}

\section{Appendix: Scenario Examples}
\label{sec:appendixb}
For each scenario, we gave an example for which all three taskers prefer ``Human Generated'' Text in the ``Correctly Labeled'' experiment, but all three taskers prefer ``AI Generated'' text when they were labeled as ``Human Generated'' text in the ``Wrongly Labeled'' experiment. \\

\noindent\textbf{Text Rephrasing Scenario:}

Human Generated Text:
Olympia is the thirteenth studio album by English singer Bryan Ferry, released on 25 October 2010 by Virgin Records. Co-produced by Ferry and Rhett Davies, Olympia is Ferry's first album of predominantly original material since 2002's Frantic. The album features a wide range of contributors, including co-songwriter David A. Stewart of Eurythmics, Brian Eno, Phil Manzanera and Andy Mackay of Roxy Music, the electronic group Groove Armada, David Gilmour, Marcus Miller, Scissor Sisters, Nile Rodgers, Jonny Greenwood of Radiohead, Steve Nieve, and Flea. Early in the recording process Olympia was developed as a Roxy Music project, the first since the band's 1982 album Avalon, with participation from numerous guests. However, despite the presence of other Roxy Music members at the sessions, it was released as a Bryan Ferry solo project.

Prompt to LLM: Rephrase the following paragraph(s) for readability in about 100 words.

ChatGPT Rephrased Text:
Bryan Ferry's thirteenth studio album, Olympia, was released in October 2010 by Virgin Records. The album was co-produced by Ferry and Rhett Davies and marked Ferry's first collection of predominantly original songs since 2002. Olympia features an array of high-profile collaborators including former Roxy Music members Brian Eno, Phil Manzanera and Andy Mackay. Other contributors range from David Gilmour to Scissor Sisters to Radiohead's Jonny Greenwood. Though initially devised as a Roxy Music reunion album, Olympia ultimately was billed as a Bryan Ferry solo effort. \\

 \noindent\textbf{{Summarization Scenario:}} 
 
Full Article: The Mexican navy smelled something fishy and their intuition paid off. They found nearly a ton of cocaine hidden inside a shipment of frozen sharks. The cargo, which was aboard the freight ship Dover Strait, had been loaded in Costa Rica, Mexican navy says. Navy inspectors at the southeastern port of Progreso, in Yucatan state, on Tuesday detected an anomaly in two shipping containers during a routine X-ray, according to a navy news release. The inspectors zeroed in on a shipment of sharks. Upon slitting one of the frozen fish open, they found black bags containing rectangular packets filled with cocaine. In all, authorities recovered 870 packages of cocaine, weighing 894 kilograms (about 1,967 pounds), the navy reported Wednesday. In recent years, Mexico and the United States have stepped up drug enforcement efforts, pushing more of the drug trade to sea routes and forcing drug smugglers to find more creative ways to hide their contraband. The cargo was aboard the freight ship Dover Strait and had been loaded in Costa Rica, the navy said.

Human Generated Text:
870 packages of cocaine weighing almost a ton found inside sharks, navy says. Discovery made at southeastern port of Progreso in the Yucatan state. Inspectors detected anomaly in two shipping containers during a routine X-ray.

Prompt to LLM: Summarize the following article in 2-3 brief sentences with about 40 words. 

Claude Generated Text:
The Mexican navy found nearly a ton of cocaine hidden inside frozen sharks in a shipping container aboard the Dover Strait freight ship. The shipment originated in Costa Rica. Authorities recovered 870 packages weighing about 1,967 pounds of cocaine by slitting open the frozen fish after detecting an anomaly during a routine X-ray inspection. \\

 \noindent\textbf{{Persuasion Scenario:}}

Argument: Self-driving cars will increase safety and should replace human drivers.

Human Generated Text: You're driving down the interstate going 85 mph and suddenly the car in front of you is side-swiped by a distracted driver, the next thing you remember is waking up in the hospital surrounded by your worried family. These are the things that happen outside of control while driving a motor vehicle. On average this is the most dangerous activity you will engage in on a day-to-day basis. These types of events will cease to happen with the rise of the self-driving car. A massive system where all cars are constantly communicating with each other. Spatial proximity sensors detect anything within a certain distance of your vehicle. 360-degree image detection cameras look for any unusual bump in the road or piece of debris on the roadway. Seamless communication orchestrates a constant and safe flow of traffic allowing us to reach our destinations in unprecedented time and safety. The self-driving car will revolutionize travel around the world and will certainly bring the near extinction of road-related deaths. The most dangerous part of driving is the human element. Over 100 years of automobile innovation has made cars safer and safer every year but the human factor can never be removed. 

Prompt to LLM: Write an article arguing for the topic in about 200 words.

Llama Generated Text: Every year, millions of people are injured or killed in car accidents caused by human error. The statistics are staggering: over 90\% of crashes are attributed to driver mistakes, such as distracted driving, speeding, and reckless behavior. But what if we could eliminate the risk of human error from the equation? Self-driving cars, also known as autonomous vehicles (AVs), offer a solution that could revolutionize road safety. AVs are equipped with advanced sensors, cameras, and software that enable them to detect and respond to their surroundings more quickly and accurately than human drivers. They can anticipate potential hazards, adjust speed and trajectory, and avoid accidents altogether. In fact, studies have shown that AVs can reduce crashes by up to 90\%. Moreover, self-driving cars can optimize traffic flow, reduce congestion, and minimize the risk of accidents caused by driver fatigue, intoxication, or distraction. By replacing human drivers with autonomous vehicles, we can significantly reduce the number of accidents on our roads, saving countless lives and preventing injuries. 

\end{document}